\documentclass[]{llncs}
\usepackage[T1]{fontenc}
\usepackage{graphicx}
\usepackage{cite}
\usepackage{amsmath,amssymb,amsfonts}
\usepackage{algorithm}
\usepackage[noend]{algorithmic}
\usepackage{textcomp}
\usepackage{lscape} 
\usepackage{booktabs}
\usepackage[colorlinks=false, hidelinks, urlcolor=black, linkcolor=black, citecolor=black]{hyperref}
\usepackage{svg}
\usepackage{tablefootnote}
\usepackage{xcolor,colortbl}
\usepackage{bm}
\usepackage[
  separate-uncertainty = true,
  multi-part-units = repeat
]{siunitx}
\usepackage{booktabs}
\usepackage{enumitem}
\usepackage{xurl}
\def\BibTeX{{\rm B\kern-.05em{\sc i\kern-.025em b}\kern-.08em
    T\kern-.1667em\lower.7ex\hbox{E}\kern-.125emX}}
\usepackage{lscape}
\begin{document}
\title{Resilience to the Flowing Unknown: an Open Set Recognition Framework for Data Streams}
%
%
\author{Marcos Barcina-Blanco\thanks{Corresponding author. Email: \email{marcos.barcina@tecnalia.com}}\inst{1,2}\orcidID{0009-0004-0278-8644} \and
Jesus L. Lobo\inst{1}\orcidID{0000-0002-6283-5148} \and
Pablo Garcia-Bringas\inst{2}\orcidID{0000-0003-3594-9534} \and
Javier Del Ser\inst{1,3}\orcidID{0000-0002-1260-9775}}
%
%
\institute{TECNALIA, Basque Research \& Technology Alliance (BRTA),\\Derio, 48160, Spain \and
Faculty of Engineering, University of Deusto, Bilbao, 48007, Spain \and
University of the Basque Country
(UPV/EHU), Bilbao, 48013, Spain}
\maketitle              
\begin{abstract}
Modern digital applications extensively integrate Artificial Intelligence models into their core systems, offering significant advantages for automated decision-making. However, these AI-based systems encounter reliability and safety challenges when handling continuously generated data streams in complex and dynamic scenarios. This work explores the concept of resilient AI systems, which must operate in the face of unexpected events, including instances that belong to patterns that have not been seen during the training process. This is an issue that regular closed-set classifiers commonly encounter in streaming scenarios, as they are designed to compulsory classify any new observation into one of the training patterns (i.e., the so-called \textit{over-occupied space} problem). In batch learning, the Open Set Recognition research area has consistently confronted this issue by requiring models to robustly uphold their classification performance when processing query instances from unknown patterns. In this context, this work investigates the application of an Open Set Recognition framework that combines classification and clustering to address the \textit{over-occupied space} problem in streaming scenarios. Specifically, we systematically devise a benchmark comprising different classification datasets with varying ratios of known to unknown classes. Experiments are presented on this benchmark to compare the performance of the proposed hybrid framework with that of individual incremental classifiers. Discussions held over the obtained results highlight situations where the proposed framework performs best, and delineate the limitations and hurdles encountered by incremental classifiers in effectively resolving the challenges posed by open-world streaming environments.

\end{abstract}

\section{Introduction} \label{Introduction}
The rapid expansion of Artificial Intelligence (AI) systems over the last decade has raised major concerns about their resilience and safety. In real-world scenarios AI-based systems are challenged by unforeseen operational conditions and novel inputs. One promising field for making AI systems resilient against Unknown Classes (UC) is the field of Open Set Recognition (OSR) \cite{geng2020recent}. Closed-set classification assumes that all query instances of the model in inference belong to one of the classes present in the training data. OSR goes beyond this assumption by confronting a case where UC may emerge during the inference phase of the model. Such an open-set scenario is common in applications that ingest streams of data \cite{fahy2022scarcity}, in which UC are more likely to occur at any point in time. In their simplest formulation, OSR models must detect instances belonging to UC, while correctly classifying those belonging to Known Classes (KC). This is essential for developing further strategies to characterize UC and to incorporate them to the knowledge base of the model.

\begin{figure}[h!]
    \centering
    \includegraphics[scale=0.7]{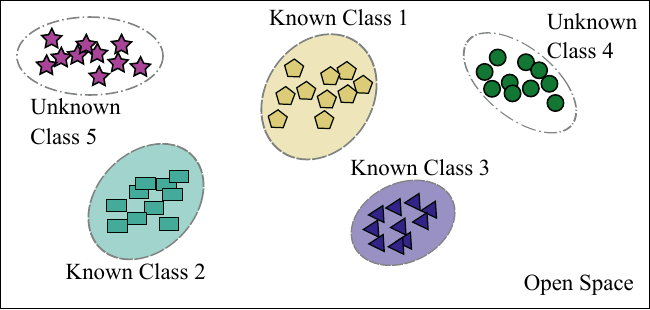}
    \caption{Example of OSR classification.}
    \label{fig:osr_img}
\end{figure}

Together with the formal definition of OSR, the seminal work in \cite{scheirer2012toward} also introduced the problem of \textit{over-occupied space}. It refers to the extra feature space that closed-set classifiers assign across all KC in order to create meaningful boundaries. Although the \textit{over-occupied space} is purposed to increase the generalization capabilities of the model, it also implies that any instance belonging to an UC will inevitably fall in this space and be incorrectly classified as a KC. Hence, traditional classification models require adaptation for open-set environments \cite{scheirer2012toward, mendes2017nearest, hui2022new}. Figure \ref{fig:osr_img} shows an example of OSR classification, where the model limits the open space assigned to the KC, and also detects samples from UC.
The recent work \cite{geng2020recent} suggests that the combination of clustering and classification can effectively address this issue. However, scarce attempts have been made hybridizing clustering and classification techniques \cite{coletta2019combining,wang2022open}. Although promising experiments in this area are confined to static batch environments, OSR scenarios are more prevalent in data streaming setups, where additional challenges exist. Models must not only deal with the appearance of UC, but also continuously adapt and learn from novel data distributions \cite{agrahari2024review}.

This work aims to showcase how a hybrid clustering and classification framework can overcome the \textit{over-occupied space} issue, evaluating its efficacy in streaming contexts with varying known-to-unknown class ratios. The main contributions are summarized as follows:
\begin{itemize}[label=\textbullet, leftmargin=*]
  \item A framework that blends together incremental clustering and classification to reduce the \textit{over-occupied space} and to yield a resilient OSR model.
  \item An evaluation of the proposed hybrid framework under streaming conditions, and a performance comparison with a regular and an incremental classifier, exposing the limitations of these single classifiers when dealing with UC.
  \item A discussion on the limitations of UC detection methods in streaming settings, and an outline of possible research directions by which clustering and classification methods can be used together to tackle OSR in non-stationary environments.
\end{itemize}

The rest of this work is structured as follows: Section \ref{Related Work} briefly revisits relevant literature on OSR, whereas Section \ref{Problem Definition} formally states the \textit{over-occupied space} problem in OSR. Next, Section \ref{Proposed Approach} provides an explanation of the proposed hybrid framework. Section \ref{Methodology} describes the datasets, methods, and metrics used in our experiments. Section \ref{Results and discussion} presents and discusses on the results from the experiments. Finally, Section \ref{Conclusion} draws concluding remarks and sketches future research paths basedon our findings.

\section{Related Work} \label{Related Work}

Before delving into the work done in OSR, it is important to mention that the appearance of UC has been tackled from other research areas in Machine Learning, such as Novelty Detection (ND) and Out-of-Distribution (OoD) detection \cite{yang2021generalized}. Although these two research areas focus on detecting and dealing with novel instances, they possess some key differences with OSR. ND is mainly considered to be an unsupervised task and does not require the classification of instances from KC. On the other hand, OoD differs in the training and benchmarking procedure. Unlike OoD, OSR avoids training with real outliers to control the positively labeled open space. While adding a "background" class is effective, it does not restrict the \textit{over-occupied space}. OSR benchmarks split a dataset into KC (\emph{in-distribution}) and UC (\emph{out-of-distribution}) instances. In contrast, OoD typically regards one dataset as in-distribution and other datasets as OoD. Contributions done so far in the field agree on the division of OSR approaches into two main categories \cite{geng2020recent,mahdavi2021survey}: discriminative and generative. We now revisit contributions in these two categories (Subsection \ref{sec:approaches}), followed by a justification of the contribution of this work (Subsection \ref{sec:contrib}).

\subsection{OSR Approaches} \label{sec:approaches}

This first category of OSR approaches (discriminative) focuses on modeling the data into smaller areas in the feature space to distinguish between KC and UC. Early approaches adapted traditional classification models to the open-world \cite{scheirer2012toward, mendes2017nearest, hui2022new}. Other approaches employ methods better suited for predicting UC, such as Gaussian Processes \cite{chen2024improving}, which assign an uncertainty measure to predictions. Deep neural networks have also been extensively used in OSR due to their strong representation capabilities \cite{cevikalp2023anomaly, komorniczak2024distance}. Alternatively, the authors in \cite{liu2024frequency} argue that representing data in the frequency domain instead of the usual spatial domain is more suited for OSR. Prototype learning tightens feature representations, enhancing boundary clarity between KC and UC \cite{din2024reliable}.

The second class of OSR approaches (generative) seeks to generate artificial instances that represent the UC and train the model with them. Most works done in this line rely on Generative Adversarial Networks (GAN) \cite{kong2021opengan}. Recent activity has placed greater emphasis on the quality of the synthesized instances \cite{pal2023morgan} and on learning extra synthesized classes to make the model reserve space suitable for future UC \cite{ma2024iosl}.

\subsection{Contribution} \label{sec:contrib}

In discriminative approaches, clustering and classification models can be combined to address the \textit{over-occupied space} problem. Clustering models adjust KC to a smaller area, while classification models are used to discriminate between them. Approaches in the field employ sequential \cite{wang2022open} or simultaneous \cite{coletta2019combining, zhang2020hybrid} strategies. Finally, the research in streaming conditions tackles challenges like identifying new classes from unknown instances and updating models with new knowledge, which still remain untackled \cite{gao2019sim,leo2020moving}. Few works explicitly address OSR in streaming setups \cite{mundt2022unified}. However, the literature lacks a detailed exploration of the \textit{over-occupied space} problem and the connection of OSR to ND and concept evolution in stream learning \cite{agrahari2024review}.

\section{Problem Statement} \label{Problem Definition}
Traditional streaming classification tasks assume that there is an infinite sequence of pairs $(\mathbf{x}_t, y_t)$, where $\mathbf{x}_t\in\bm{\mathcal{X}}$ and $y_t \in \mathcal{C}_{KC} = \{y_1, y_2, \ldots, y_N\}$ (KC). In the absence of unexpected classes and non-stationarities, classifiers are either trained offline with instances from $\mathcal{C}_{KC}$ or, alternatively, learn incrementally from the stream if no extreme verification latency is held. Classifiers compute a probability distribution over the training classes $p(y|\mathbf{x})$ with $y \in \mathcal{C}_{KC}$. Since this conditional distribution is defined for any $\mathbf{x}\in\bm{\mathcal{X}}$, feature space is over-occupied by the classifier as it characterizes regions in $\bm{\mathcal{X}}$ that may eventually belong to UC. Since any predicted label $\hat{y}_t$ belongs to $\mathcal{C}_{KC}$, whenever a new query instance $\mathbf{x}_t$ whose true label does not belong to $\mathcal{C}_{KC}$ appears in the stream, a classification error occurs.

By contrast, streaming OSR classification models are designed to work in an environment similar to the above, but where $y_t$ does not necessarily belong to $\mathcal{C}_{KC}$ $\forall t$. They are trained to correctly classify instances $\mathbf{x}_t$ from $\mathcal{C}_{KC}$, but instances may also belong to $\mathcal{C}_{UC}$, with $\mathcal{C}_{KC}\cap\mathcal{C}_{UC}=\emptyset$. A streaming OSR model should be able to determine whether any new instance $\mathbf{x}_t$ belongs to $\mathcal{C}_{KC}$ or $\mathcal{C}_{UC}$. Additionally, if $\mathbf{x}_t$ is determined to be from $\mathcal{C}_{KC}$, the model should also predict its class label $\hat{y}_t \in \mathcal{C}_{KC}$.

\section{Proposed Streaming OSR Framework} \label{Proposed Approach}

As explained in the previous section, the predictions of a single closed-set classifier are limited to $C_{KC}$. The addition of a clustering model can help circumvent this flaw by capturing the structure of arriving data, whereas a closed-set classification model can be deployed and updated incrementally with instances arriving from the stream verified to belong to $\mathcal{C}_{KC}$. 
\begin{figure}[h!]
    \vspace{-1mm}
    \centering
    \includegraphics[scale=0.6]{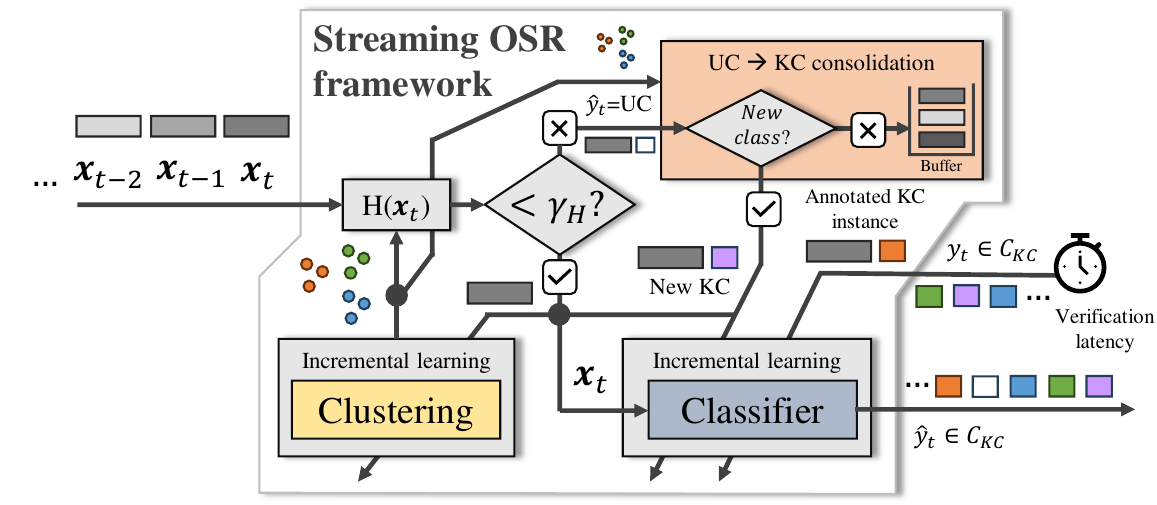}
    \caption{Block diagram of the proposed streaming OSR framework.}
    \label{fig:framework}
\end{figure}

This rationale guides the design of the proposed OSR framework for data streams depicted in Figure \ref{fig:framework}. Stream instances $\{\mathbf{x}_t\}$ are received over time, each first processed in order to estimate the degree to which stream instance $\mathbf{x}_t$ is unknown. 
This estimation exploits an incremental clustering algorithm to characterize the feature space, continuously updating cluster arrangements upon each arrival of instances. Let $M_t$ denote the number of clusters discovered by the cluster algorithm at time $t$, and $\mathbf{c}_t^m\in\bm{\mathcal{X}}$ the centroid corresponding to the $m$-th cluster. A measure of novelty of $\mathbf{x}_t$ can be produced through the use of entropy \cite{shannon1948mathematical,meyer2019importance}. Based on the assumption that instances closer to each other are more similar than those further apart, we can compute a pseudo-probability $\tilde{p}(m|\mathbf{x}_t)$ of every arriving instance $\mathbf{x}_t$ to every cluster $m\in\{1,\ldots,M_t\}$:
\begin{equation} \label{eq:one}
\tilde{p}(m|\mathbf{x}_t) = \frac{1/d(\mathbf{x}_t, \mathbf{c}_t^{m})^2}{\sum_{m'=1}^{M_t} 1/d(\mathbf{x}_t, \mathbf{c}_t^{m'})^2}
\end{equation}
, where $d(\cdot,\cdot)$ stands for the Euclidean distance. The $1/d(\mathbf{x}_t, \mathbf{c}_t^{m})^2$ is adopted as the weight representation function because it quickly diminishes with distance, granting that those elements nearer to the clusters show much higher weights than those farther away. Based on the pseudo-probability vector $\{\tilde{p}(m|\mathbf{x}_t)\}_{m=1}^{M_t}$, the framework computes the entropy associated to $\mathbf{x}_t$ as:
\begin{equation}\label{eq:H}
H(\mathbf{x}_t) = \sum\limits_{m=1}^{M_t} -\tilde{p}(m|\mathbf{x}_t)\log_{M_t} \tilde{p}(m|\mathbf{x}_t)
\end{equation}
, in which instances with lower $H(\mathbf{x}_t)$ belong to $\mathcal{C}_{KC}$, while instances with higher entropy are declared to be unknown.
\begin{algorithm}[hbt!]
\caption{Proposed streaming OSR framework}\label{alg:one}
\begin{algorithmic}[1]
\REQUIRE $\mathcal{C}_{KC}$, incremental clustering algorithm $f_\odot(\mathbf{x})$, incremental classifier $f_C(\mathbf{x})$, threshold $\gamma_H$
\ENSURE Warm-up with KC instances providing a classifier $f_C(\mathbf{x})$ and centroids $\{\mathbf{c}_1^m\}_{m=1}^{M_1}$ via $f_\odot(\mathbf{x})$ at $t=1$
\FORALL{$t\in\{1,\ldots,\infty\}$}
    \STATE Collect $\mathbf{x}_t\in \bm{\mathcal{X}}$ arriving from the stream
    \STATE Compute $\tilde{p}(m|\mathbf{x}_t)$ $\forall m\in\{1,\ldots,M_t\}$ as per Eq. \eqref{eq:one}
    \STATE Compute $H(\mathbf{x}_t)$ as per Eq. \eqref{eq:H}
    \IF{$H(\mathbf{x}_t)<\gamma_{H}$}
        \STATE Predict $\hat{y}_t=f_C(\mathbf{x}_t)=\arg \max_{y\in\mathcal{C}_{KC}} p(y|\mathbf{x}_t)$
        \STATE Update $f_C(\mathbf{x}_t)$ with $(\mathbf{x}_t,y_t)$ ({\small{s.t. verification latency}})
        \STATE Update $\{\mathbf{c}_t^m\}_{m=1}^{M_t}$ to $\{\mathbf{c}_{t+1}^m\}_{m=1}^{M_{t+1}}$ via $f_{\odot}(\mathbf{x}_t)$
    \ELSE
        \STATE Analyze $\mathbf{x}_t$ and other buffered UC instances 
        \IF{criterion to consolidate a new class $y^{\ast}$ is met}
            \STATE Predict $\hat{y}_t=y^{\ast}$
            \STATE Update $f_C(\mathbf{x}_t)$ with $(\mathbf{x}_t,y^{\ast})$
            \STATE Update $\{\mathbf{c}_t^m\}_{m=1}^{M_t}$ to $\{\mathbf{c}_{t+1}^m\}_{m=1}^{M_{t+1}}$ via $f_{\odot}(\mathbf{x}_t)$
        \ELSE
            \STATE Predict $\hat{y}_t=\text{UC}$ (\emph{unknown})    
        \ENDIF
    \ENDIF    
\ENDFOR
\end{algorithmic}
\end{algorithm}

The workflow is summarized in Algorithm \ref{alg:one}. New instances first arrive to the clustering model, where their entropy is measured against a predefined threshold $\gamma_H\in\mathbb{R}(0,1]$. Instances with $H(\mathbf{x}_t)\geq\gamma_H$ are considered \emph{unknown}, so they are predicted as UC and then, fed to a UC$\rightarrow$KC consolidation module. Based on the current UC instance and the current cluster space, this module decides whether a new KC should be incrementally learned by the classifier. Diverse criteria can be followed for this purpose based on a buffer of recently detected UC instances. If the presence of a new KC is determined, the instance $\mathbf{x}_t$ annotated with a new class identifier $y^{\ast}$ is fed and incrementally learned by the classifier and the clustering algorithm. By contrast, instances with $H(\mathbf{x}_t)<\gamma_H$ are passed to the clustering algorithm and the classifier for normal closed-set classification, yielding prediction $\hat{y}_t\in\mathcal{C}_{KC}$. KC samples are retained until their true label is known, whose feasibility depends on the buffer size requirements imposed by the verification latency of the stream. If feasible, the classifier can be updated incrementally with each stored $\mathbf{x}_t$ and its annotation $y_t$.
\section{Experimental Setup} \label{Methodology}

Before delving into the limitations of the proposed framework and forthcoming challenges, we review the results of an experimental benchmark. We aim to empirically assess the performance of the framework over several streaming datasets under different ratios of known-to-unknown classes. Furthermore, we consider three baselines:
\begin{enumerate}[leftmargin=*]
\item A regular, non-incremental classifier that does not update its knowledge nor is it helped by a clustering algorithm to determine whether stream instances are known (\texttt{static}).
\item An incremental classifier updated with the verified labels of the incoming stream instance, but without any functionality to detect unknown instances (\texttt{incremental}).
\item The proposed stream OSR framework (\texttt{sOSR}).
\end{enumerate}

As will be later discussed, the UC$\rightarrow$KC consolidation module can be affected by the presence of non-stationarities in the classification task (\emph{concept drift} \cite{sato2021survey}). To avoid a comparison biased by the design of this module, we follow a common practice in the literature and assume a \emph{test-then-train} scheme, by which the true label $y_t$ of $\mathbf{x}_t$ is made known to \texttt{incremental} and \texttt{sOSR} immediately after predicting $\hat{y}_t$. Given the immediate verification of every test instance, the consolidation module in the \texttt{sOSR} framework is not needed. All UC instances will be grouped into one UC to focus on the detection ability of this unified UC by the proposed framework. Works that address this problem usually apply clustering algorithms on the UC samples in the data buffer to extract new classes \cite{din2024reliable}.
\begin{table}[ht!]
\centering
\vspace{-3mm}
\caption{Parameters from the synthetic datasets}
\label{tab:dataset_parameters} 
\resizebox{0.75\columnwidth}{!}{\begin{tabular}{@{}ccccccc@{}}
\toprule
Parameters & D1-D4 & D5-D8 & D9-11 & D12-D14 & D15-17 & D18-D20 \\ \midrule
Number of instances & 1000 & 2000 & 3000 & 5000 & 7500 & 10000 \\
Number of classes & 5 & 8 & 10 & 12 & 15 & 20 \\
Number of features & 3 & 4 & 6 & 8 & 10 & 12 \\
StdDev (\texttt{isoGauss}) & 0.75 & 1 & 1.25 & 1.5 & 1.75 & 2 \\
Class Sep. (\texttt{hyperCube}) & 1.4 & 1.2 & 1 & 0.8 & 0.6 & 0.4 \\ \bottomrule
\multicolumn{7}{l}{StdDev: intra-cluster standard deviation. Class Sep.: factor} \\
\multicolumn{7}{l}{multiplying the hypercube size.}
\end{tabular}}
\end{table}

We evaluate the above baselines over $40$ synthetic datasets and one real dataset (\texttt{insects} \cite{souza2020challenges}) at different values of missing classes $\beta=|\mathcal{C}_{UC}|/(|\mathcal{C}_{KC}|+|\mathcal{C}_{UC}|)$. For example, a dataset with $10$ classes are evaluated at $\beta=0.2$ and are composed of $|\mathcal{C}_{KC}|=8$ KC during training, while $|\mathcal{C}_{KC}|+|\mathcal{C}_{UC}|=10$ classes may appear during inference, with $|\mathcal{C}_{UC}|=2$ UC. KC data instances are split into training ($80\%$) and testing ($20\%$) partitions. We then interleave at random test data instances of the KC with a $10\%$ of the total instances from each UC. Synthetic datasets consist of $20$ isotropic Gaussian point clouds (\texttt{isoGauss}) and $20$ classification datasets generated by an multidimensional adaptation of the hypercube-based Medelon dataset generation procedure in \cite{Guyon2003DesignOE} (\texttt{hyperCube}). Table \ref{tab:dataset_parameters} describes the different parameters of each synthetic dataset, from the simplest ($D_1$-$D_4$) to the most challenging ones ($D_{18}$-$D_{20}$). The \texttt{insects} dataset poses a multi-class classification problem with $52,848$ instances across $6$ types of insects, each described by $33$ features.

For the classification models, we make use of the implementation of Softmax classification, which suffers from the \textit{over-occupied space} problem. For the clustering model, we choose STREAMKmeans, which is an adaptation of the K-means method. Both are provided by the River framework \cite{montiel2021river} which is widely used by the scientific community focused on stream learning. The STREAMKmeans algorithm requires to set beforehand the number of clusters to run. Since a warm-up period is assumed (Algorithm \ref{alg:one}), we use the number of KC for this purpose. For a fair comparison, we do not perform hyperparameter tuning on any model. Models are configured with their default values from the River framework as the search of the best individual performance is not the focus of this work.

We evaluate the implemented models based on KC (KC-Acc) and UC (UC-Acc) accuracy. We adopt the open macro F1 metric (F1) defined in \cite{mendes2017nearest} which is similar to the conventional macro F1-measure, but true positives from UC are not considered. We also use the Area Under the ROC Curve (AUROC) to evaluate the baselines on the reduction of \textit{over-occupied space}. Unlike other metrics, AUROC does not need threshold a selection. However, it is not applicable to single classifier models as they cannot explicitly predict UC. As such, we do not report the AUROC associated to \texttt{static} and \texttt{incremental} baselines. The true positive rates and the false positive rates used to calculate the AUROC are sorted by their Youden Index \cite{youden1950index}. The threshold with the highest Youden Index is selected for evaluating the proposed \texttt{sOSR}. We report on average results of each baseline across all datasets for every value of $\beta$. To assess the significance of performance gaps, a non-parametric Wilcoxon signed-rank test with significance level $\alpha=0.05$ is performed over the scores of the \texttt{static} and \texttt{incremental} baselines with the scores of the \texttt{sOSR} baseline for every $\beta$. For the sake of transparency and to support follow-up studies, all scripts, datasets and results have been made publicly available in a GitHub repository\footnote{\url{https://github.com/Unknownhedgehog/CCH_overoccupied_space_OSR}}.

\section{Results and Discussion} \label{Results and discussion}

\begin{table*}[ht]
\caption{\label{tab:performance_comparison} Results$^{\mathrm{a}}$ (average$\pm$std) over synthetic datasets.}
\centering
\resizebox{1\columnwidth}{!}{\begin{tabular}{@{}ccccclcccc@{}}
\toprule
 & \multicolumn{4}{c}{\texttt{isoGauss}} &  & \multicolumn{4}{c}{\texttt{hyperCube}} \\ \cmidrule(lr){2-5} \cmidrule(l){7-10} 
$\beta$ & KC-Acc & UC-Acc & F1 & AUC &  & KC-Acc & UC-Acc & F1 & AUC \\ \midrule
0.1 & \begin{tabular}[c]{@{}c@{}}\cellcolor[gray]{0.8}$0.98 \pm 0.05$\\\cellcolor[gray]{0.8}$0.98 \pm 0.04$ \\ $0.59 \pm 0.15$\end{tabular} & \begin{tabular}[c]{@{}c@{}}$0.00 \pm 0.00$\\ $0.04 \pm 0.09$\\ \cellcolor[gray]{0.8}$0.92 \pm 0.05$\end{tabular} & \begin{tabular}[c]{@{}c@{}}\cellcolor[gray]{0.8}$0.84 \pm 0.08$\\ \cellcolor[gray]{0.8}$0.84 \pm 0.08$\\ $0.55 \pm 0.11$\end{tabular} & \begin{tabular}[c]{@{}c@{}}-\\ -\\ $0.73 \pm 0.11$\end{tabular} &  & \begin{tabular}[c]{@{}c@{}}\cellcolor[gray]{0.8}$0.55 \pm 0.25$\\\cellcolor[gray]{0.8}$0.55 \pm 0.25$ \\ $0.28 \pm 0.2$\end{tabular} & \begin{tabular}[c]{@{}c@{}}$0.00 \pm 0.00$\\ $0.00 \pm 0.00$\\ \cellcolor[gray]{0.8}$0.75 \pm 0.08$\end{tabular} & \begin{tabular}[c]{@{}c@{}}\cellcolor[gray]{0.8}$0.46 \pm 0.18$\\ \cellcolor[gray]{0.8}$0.46 \pm 0.18$\\ $0.28 \pm 0.17$\end{tabular} & \begin{tabular}[c]{@{}c@{}}-\\ -\\ $0.59 \pm 0.10$\end{tabular} \\
\midrule
0.25 & \begin{tabular}[c]{@{}c@{}}\cellcolor[gray]{0.8}$0.98 \pm 0.04$\\ \cellcolor[gray]{0.8}$0.98 \pm 0.03$\\ $0.66 \pm 0.13$\end{tabular} & \begin{tabular}[c]{@{}c@{}}$0.00 \pm 0.00$\\ $0.03 \pm 0.09$\\ \cellcolor[gray]{0.8}$0.90 \pm 0.07$\end{tabular} & \begin{tabular}[c]{@{}c@{}}\cellcolor[gray]{0.8}$0.83 \pm 0.08$\\ \cellcolor[gray]{0.8}$0.84 \pm 0.07$\\ $0.61 \pm 0.08$\end{tabular} & \begin{tabular}[c]{@{}c@{}}-\\ -\\ $0.80 \pm 0.09$\end{tabular} &  & \begin{tabular}[c]{@{}c@{}}\cellcolor[gray]{0.8}$0.57 \pm 0.24$\\ \cellcolor[gray]{0.8}$0.57 \pm 0.24$\\ $0.29 \pm 0.18$\end{tabular} & \begin{tabular}[c]{@{}c@{}}$0.00 \pm 0.00$\\ $0.00 \pm 0.01$\\ \cellcolor[gray]{0.8}$0.74 \pm 0.08$\end{tabular} & \begin{tabular}[c]{@{}c@{}}\cellcolor[gray]{0.8}$0.47 \pm 0.18$\\ \cellcolor[gray]{0.8}$0.47 \pm 0.18$\\ $0.29 \pm 0.16$\end{tabular} & \begin{tabular}[c]{@{}c@{}}-\\ -\\ $0.58 \pm 0.07$\end{tabular} \\
\midrule
0.4 & \begin{tabular}[c]{@{}c@{}}\cellcolor[gray]{0.8}$0.99 \pm 0.02$\\ \cellcolor[gray]{0.8}$0.99 \pm 0.02$\\ $0.75 \pm 0.13$\end{tabular} & \begin{tabular}[c]{@{}c@{}}$0.00 \pm 0.00$\\ $0.05 \pm 0.10$\\ \cellcolor[gray]{0.8}$0.91 \pm 0.05$\end{tabular} & \begin{tabular}[c]{@{}c@{}}\cellcolor[gray]{0.8}$0.81 \pm 0.07$\\ \cellcolor[gray]{0.8}$0.81 \pm 0.07$\\ $0.66 \pm 0.07$\end{tabular} & \begin{tabular}[c]{@{}c@{}}-\\ -\\ $0.86 \pm 0.08$\end{tabular} &  & \begin{tabular}[c]{@{}c@{}}\cellcolor[gray]{0.8}$0.62 \pm 0.24$\\ \cellcolor[gray]{0.8}$0.62 \pm 0.24$\\ $0.33 \pm 0.20$\end{tabular} & \begin{tabular}[c]{@{}c@{}}$0.00 \pm 0.00$\\ $0.03 \pm 0.11$\\\cellcolor[gray]{0.8}$0.74 \pm 0.08$\end{tabular} & \begin{tabular}[c]{@{}c@{}}\cellcolor[gray]{0.8}$0.49 \pm 0.16$\\ \cellcolor[gray]{0.8}$0.49 \pm 0.16$\\ $0.32 \pm 0.16$\end{tabular} & \begin{tabular}[c]{@{}c@{}}-\\ -\\ $0.61 \pm 0.11$\end{tabular} \\
\midrule
0.6 & \begin{tabular}[c]{@{}c@{}}\cellcolor[gray]{0.8}$1.00 \pm 0.01$\\ \cellcolor[gray]{0.8}$1.00 \pm 0.01$\\ $0.84 \pm 0.11$\end{tabular} & \begin{tabular}[c]{@{}c@{}}$0.00 \pm 0.00$\\ $0.24 \pm 0.24$\\ \cellcolor[gray]{0.8}$0.93 \pm 0.05$\end{tabular} & \begin{tabular}[c]{@{}c@{}}\cellcolor[gray]{0.8}$0.74 \pm 0.09$\\ \cellcolor[gray]{0.8}$0.75 \pm 0.08$\\ $0.68 \pm 0.05$\end{tabular} & \begin{tabular}[c]{@{}c@{}}-\\ -\\ $0.92 \pm 0.06$\end{tabular} &  & \begin{tabular}[c]{@{}c@{}}\cellcolor[gray]{0.8}$0.70 \pm 0.24$\\ \cellcolor[gray]{0.8}$0.69 \pm 0.24$\\ $0.37 \pm 0.17$\end{tabular} & \begin{tabular}[c]{@{}c@{}}$0.00 \pm 0.00$\\ $0.11 \pm 0.19$\\ \cellcolor[gray]{0.8}$0.74 \pm 0.10$\end{tabular} & \begin{tabular}[c]{@{}c@{}}\cellcolor[gray]{0.8}$0.50 \pm 0.13$\\ \cellcolor[gray]{0.8}$0.50 \pm 0.13$\\ $0.33 \pm 0.13$\end{tabular} & \begin{tabular}[c]{@{}c@{}}-\\ -\\ $0.61 \pm 0.08$\end{tabular} \\
\midrule
0.75 & \begin{tabular}[c]{@{}c@{}}\cellcolor[gray]{0.8}$1.00 \pm 0.00$\\ \cellcolor[gray]{0.8}$1.00 \pm 0.01$\\ $0.90 \pm 0.07$\end{tabular} & \begin{tabular}[c]{@{}c@{}}$0.00 \pm 0.00$\\ $0.39 \pm 0.23$\\ \cellcolor[gray]{0.8}$0.95 \pm 0.04$\end{tabular} & \begin{tabular}[c]{@{}c@{}}$0.66 \pm 0.08$\\ $0.69 \pm 0.07$\\ $0.67 \pm 0.04$\end{tabular} & \begin{tabular}[c]{@{}c@{}}-\\ -\\ $0.96 \pm 0.03$\end{tabular} &  & \begin{tabular}[c]{@{}c@{}}\cellcolor[gray]{0.8}$0.79 \pm 0.21$\\ \cellcolor[gray]{0.8}$0.78 \pm 0.20$\\ $0.39 \pm 0.19$\end{tabular} & \begin{tabular}[c]{@{}c@{}}$0.00 \pm 0.00$\\ $0.16 \pm 0.17$\\ \cellcolor[gray]{0.8}$0.81 \pm 0.07$\end{tabular} & \begin{tabular}[c]{@{}c@{}}\cellcolor[gray]{0.8}$0.51 \pm 0.09$\\ \cellcolor[gray]{0.8}$0.51 \pm 0.01$\\ $0.33 \pm 0.12$\end{tabular} & \begin{tabular}[c]{@{}c@{}}-\\ -\\ $0.63 \pm 0.01$\end{tabular} \\ \bottomrule
\multicolumn{10}{l}{$^{\mathrm{a}}$The upper result in each cell corresponds to the \texttt{static} classifier, the middle result corresponds} \\
\multicolumn{10}{l}{to the \texttt{incremental} classifier, and the bottom result stands for the scores of the proposed \texttt{sOSR}} \\
\multicolumn{10}{l}{framework.}
\end{tabular}}
\vspace{-3mm}
\end{table*}
The results of the experiments are shown in Tables \ref{tab:performance_comparison} (synthetic datasets) and \ref{tab:performance_comparison2} (real dataset) as average $\pm$ standard deviation for the three baselines: \texttt{static} (top position in each cell), \texttt{incremental} (middle) and \texttt{sOSR} (bottom). Best results for every ($\beta$, dataset, score) combination considering the statistical significance as per the Wilcoxon test are highlighted in gray. As expected, the regular \texttt{static} classifier lacks the ability to detect any instances from the UC. The UC-Acc of the \texttt{incremental} classifier grows with higher levels of missing classes and higher number of instances from the UC. The KC-Acc and F1 are generally higher for the single classifier methods. Overall, the UC-Acc is the highest for the proposed \texttt{sOSR} approach, as it is able to better detect instances from UC, and to manage the \textit{over-occupied space} than the other classifiers. \begin{table}[H]
\vspace{-1mm}
\centering
\caption{\label{tab:performance_comparison2} Results$^{\mathrm{a}}$ (average$\pm$std) over the \texttt{insects} dataset.}
\resizebox{0.65\columnwidth}{!}{\begin{tabular}{@{}cccccl@{}}
\toprule
$\beta$ & KC-Acc & UC-Acc & F1 & AUC &  \\ \midrule
0.1 & \begin{tabular}[c]{@{}c@{}}\cellcolor[gray]{0.8}$0.67 \pm 0.01$\\ \cellcolor[gray]{0.8}$0.65 \pm 0.00$\\ $0.35 \pm 0.15$\end{tabular} & \begin{tabular}[c]{@{}c@{}}$0.00 \pm 0.00$\\ $0.14 \pm 0.02$\\ \cellcolor[gray]{0.8}$0.62 \pm 0.20$\end{tabular} & \begin{tabular}[c]{@{}c@{}}\cellcolor[gray]{0.8}$0.52 \pm 0.01$\\ \cellcolor[gray]{0.8}$0.52 \pm 0.00$\\ $0.34 \pm 0.11$\end{tabular} & \begin{tabular}[c]{@{}c@{}}-\\ -\\ $0.57 \pm 0.04$\end{tabular} &  \\
\midrule
0.25 & \begin{tabular}[c]{@{}c@{}}\cellcolor[gray]{0.8}$0.64 \pm 0.01$\\ \cellcolor[gray]{0.8}$0.62 \pm 0.01$\\ $0.41 \pm 0.15$\end{tabular} & \begin{tabular}[c]{@{}c@{}}$0.00 \pm 0.00$\\ $0.31 \pm 0.02$\\ \cellcolor[gray]{0.8}$0.54 \pm 0.17$\end{tabular} & \begin{tabular}[c]{@{}c@{}}\cellcolor[gray]{0.8}$0.47 \pm 0.01$\\ \cellcolor[gray]{0.8}$0.47 \pm 0.00$\\ $0.36 \pm 0.12$\end{tabular} & \begin{tabular}[c]{@{}c@{}}-\\ -\\ $0.55 \pm 0.05$\end{tabular} &  \\
\midrule
0.5 & \begin{tabular}[c]{@{}c@{}}\cellcolor[gray]{0.8}$0.73 \pm 0.01$\\ \cellcolor[gray]{0.8}$0.70 \pm 0.00$\\ $0.49 \pm 0.05$\end{tabular} & \begin{tabular}[c]{@{}c@{}}$0.00 \pm 0.00$\\ $0.51 \pm 0.02$\\ \cellcolor[gray]{0.8}$0.69 \pm 0.04$\end{tabular} & \begin{tabular}[c]{@{}c@{}}\cellcolor[gray]{0.8}$0.51 \pm 0.01$\\ \cellcolor[gray]{0.8}$0.51 \pm 0.00$\\ $0.42 \pm 0.03$\end{tabular} & \begin{tabular}[c]{@{}c@{}}-\\ -\\ $0.64 \pm 0.04$\end{tabular} &  \\
\midrule
0.7 & \begin{tabular}[c]{@{}c@{}}\cellcolor[gray]{0.8}$0.96 \pm 0.00$\\ \cellcolor[gray]{0.8}$0.93 \pm 0.00$\\ $0.59 \pm 0.21$\end{tabular} & \begin{tabular}[c]{@{}c@{}}$0.00 \pm 0.00$\\ $0.14 \pm 0.02$\\ \cellcolor[gray]{0.8}$0.47 \pm 0.17$\end{tabular} & \begin{tabular}[c]{@{}c@{}}\cellcolor[gray]{0.8}$0.57 \pm 0.00$\\ \cellcolor[gray]{0.8}$0.57 \pm 0.00$\\ $0.43 \pm 0.14$\end{tabular} & \begin{tabular}[c]{@{}c@{}}-\\ -\\ $0.52 \pm 0.02$\end{tabular} &  \\ \bottomrule
\multicolumn{5}{l}{$^{\mathrm{a}}$Same interpretation of scores in each cell as in Table \ref{tab:performance_comparison}.}
\end{tabular}}
\vspace{-5mm}
\end{table}Furthermore, the gaps between baselines in terms of the KC-Acc and UC-Acc scores are statistically significant. The results of F1 are an exception for higher levels of missing classes due to a more balanced amount of instances from KC and UC. The higher F1 scores of single classifier approaches (\texttt{static} and \texttt{incremental}) compared to the proposed \texttt{sOSR} arise from the majority of arriving instances being KC. This trend is partially replicated by the \texttt{incremental} classifier. Notably, the  performance of the \texttt{incremental} classifier when detecting UC instances varies greatly with levels of missing classes. Initially, with few UC instances, it struggles to learn a new space for the UC, reflected in extremely low UC-Acc scores at lower $\beta$ that gradually improve with higher $\beta$. Additionally, as UC detection improves, KC accuracy tends to decrease. Moreover, the standard deviation of UC-Acc at $\beta\in\{0.6,0.75\}$ also becomes higher due to the unpredictability of the position of the UC in the feature space. Although the \texttt{incremental} classifier is learning a new space for the new instances, it may be at the expense of the space occupied by the KC, which does not solve the \textit{over-occupied space} problem robustly. 

\section{Conclusions, Limitations and Future Work} \label{Conclusion}

This work has elaborated on the \textit{over-occupied space} issue of OSR, focusing on its complexity in streaming environments. We have proposed a streaming OSR framework, that combines an incremental classifier with incremental clustering, to yield a generic processing workflow to perform classification modeling tasks over open-world data streams. We have compared three approaches: a regular classifier, an incremental classifier, and the streaming OSR framework configured with an incremental version of the K-Means clustering. Our results have exposed that the proposed framework can help a classifier overcome the \textit{over-occupied space} problem by virtue of a simple entropy-based detection based on the incrementally updated distribution of clusters. 

\begin{figure}[ht!]
\vspace{-2mm}
\centering
\includegraphics[scale=0.5]{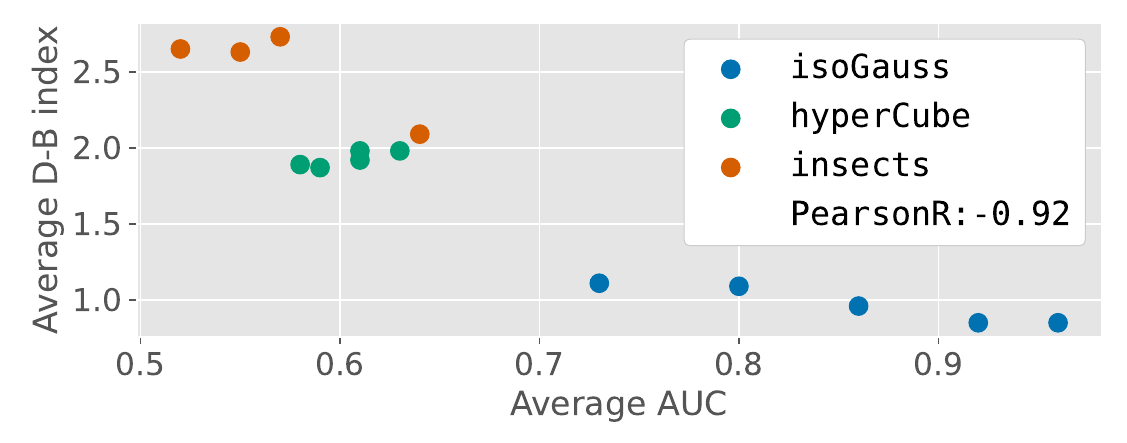}
\caption{Average AUC vs D-B index for the \texttt{sOSR} approach and the datasets from the benchmark at different values of $\beta$.}

\label{fig:davies_bouldin_comparison}
\end{figure}

Although our proposal effectively learns to detect UC, it still undergoes similar issues to a regular classifier and is not enough to solve the issue.
As discussed in Section \ref{Proposed Approach}, the reliability of the \textit{entropy} measure depends on the ability of the clustering model to find representative centers. This statement is buttressed by Figure \ref{fig:davies_bouldin_comparison}, which depicts the Davies-Bouldin (D-B) index of the \texttt{sOSR} approach for all datasets at each level of $\beta$. The higher the D-B index is, the lower the quality of the clustering will be. The core limitation of the STREAMKmeans algorithm, together with the fixed value of $\gamma_H$ used in the experiments, make the overall framework not to be suitable to properly characterize all data streams in the benchmark, as evinced by the correlation between the D-B indices and the AUC scores of the \texttt{sOSR} framework in Tables \ref{tab:performance_comparison} and \ref{tab:performance_comparison2}. An incorrect threshold also damages the performance of the closed-set classifier due to many KC instances being identified as UC. Future work will explore the use of automated configuration techniques and other incremental clustering algorithms, potentially better suited for complex streaming data distributions. 

Another aspect demanding further thought is the distinction between the different types of concept drift and OSR. Unlike virtual ($p_\mathbf{X}(\mathbf{x})$ changes) and real drift ($p_{Y|\mathbf{X}}(y|\mathbf{x})$ changes), in OSR what changes is the discrete support $\mathcal{Y}$ of the statistical variable $Y$, which represents the different classes in the stream. It can be challenging for the UC$\rightarrow$KC consolidation module to decide whether a set of UC instances detected over time corresponds to a sharp distributional change of $\mathbf{X}$ (i.e. a virtual drift) or, alternatively, should be consolidated as a new class and incorporated as such to the classifier. These limitations and other points of improvement will guide our research efforts in the near future, with the overarching goal of discovering hybrid streaming OSR approaches that operate resiliently with evolving data in open-world environments.

\subsubsection{Acknowledgements}
All authors approved the version of the manuscript to be published. This research has received funding from the Basque Government (BEREZ-IA project with grant number KK-2023/00012), and the research groups MATHMODE (IT1456-22) and D4K-Deusto for Knowledge (IT1528-22).

%
%
%
\bibliographystyle{splncs04}
\bibliography{references}
\end{document}